\documentclass[letterpaper]{article} 
\usepackage{aaai24}  
\usepackage{times}  
\usepackage{helvet}  
\usepackage{courier}  
\usepackage[hyphens]{url}  
\usepackage{graphicx} 
\usepackage{natbib}  
\usepackage{multirow}
\usepackage{subfigure}
\usepackage{xcolor}
\usepackage{amsthm,amsmath,amssymb}
\usepackage{caption} 
\usepackage{algorithm}
\usepackage{algorithmic}

\usepackage{newfloat}
\usepackage{listings}
\DeclareCaptionStyle{ruled}{labelfont=normalfont,labelsep=colon,strut=off} 
\floatstyle{ruled}
\newfloat{listing}{tb}{lst}{}
\floatname{listing}{Listing}
\nocopyright

\newcommand{\bp}[1]{\textcolor{black}{#1}}
\newcommand{\yep}[1]{\textcolor{black}{#1}}
\newcommand{\wl}[1]{\textcolor{black}{#1}}
\newcommand{\tsj}[1]{\textcolor{black}{#1}}
\newcommand{\tsjadd}[1]{\textcolor{black}{#1}}
\newcommand{\tsjaddd}[1]{\textcolor{black}{#1}}
\newcommand{\lwh}[1]{\textcolor{black}{#1}}
\newcommand{\ct}[1]{\textcolor{black}{#1}}
\title{Boosting Residual Networks with Group Knowledge}
\author{
    Shengji Tang\textsuperscript{\rm 1}\equalcontrib, Peng Ye\textsuperscript{\rm 1}\equalcontrib, Baopu Li,\\ Weihao Lin\textsuperscript{\rm 1}, Tao Chen\textsuperscript{\rm 1}\thanks{Corresponding author}, Tong He\textsuperscript{\rm 3}, Chong Yu\textsuperscript{\rm 2}, Wanli Ouyang\textsuperscript{\rm 3}
}
\affiliations{
    \textsuperscript{\rm 1} School of Information
Science and Technology, Fudan University, Shanghai, China\\

    \textsuperscript{\rm 2} Academy for Engineering and Technology, Fudan University, Shanghai, China\\
    \textsuperscript{\rm 3}Shanghai AI Laboratory, Shanghai, China\\
    eetchen@fudan.edu.cn
}

\usepackage{bibentry}

\begin{document}

\maketitle

\begin{abstract}
Recent research understands residual networks from a new perspective of the implicit ensemble model. From this view, previous methods such as stochastic depth and stimulative training have further improved the performance of residual networks by sampling and training of its subnets. However, they both use the same supervision for all subnets of different capacities and neglect the valuable knowledge generated by subnets during training. In this paper, we mitigate the significant knowledge distillation gap caused by using the same kind of supervision and advocate leveraging the subnets to provide diverse knowledge. Based on this motivation, we propose a group knowledge based training framework for boosting the performance of residual networks. Specifically, we implicitly divide all subnets into hierarchical groups by subnet-in-subnet sampling, aggregate the knowledge of different subnets in each group during training, and exploit upper-level group knowledge to supervise lower-level subnet group. Meanwhile, we also develop a subnet sampling strategy that naturally samples larger subnets, which are found to be more helpful than smaller subnets in boosting performance for hierarchical groups. Compared with typical subnet training and other methods, our method achieves the best efficiency and performance trade-offs on multiple datasets and network structures. The code is at https://github.com/tsj-001/AAAI24-GKT. 
\end{abstract}

\section{Introduction}

\begin{figure}[t]
  \centering
  \includegraphics[width=0.95\linewidth]{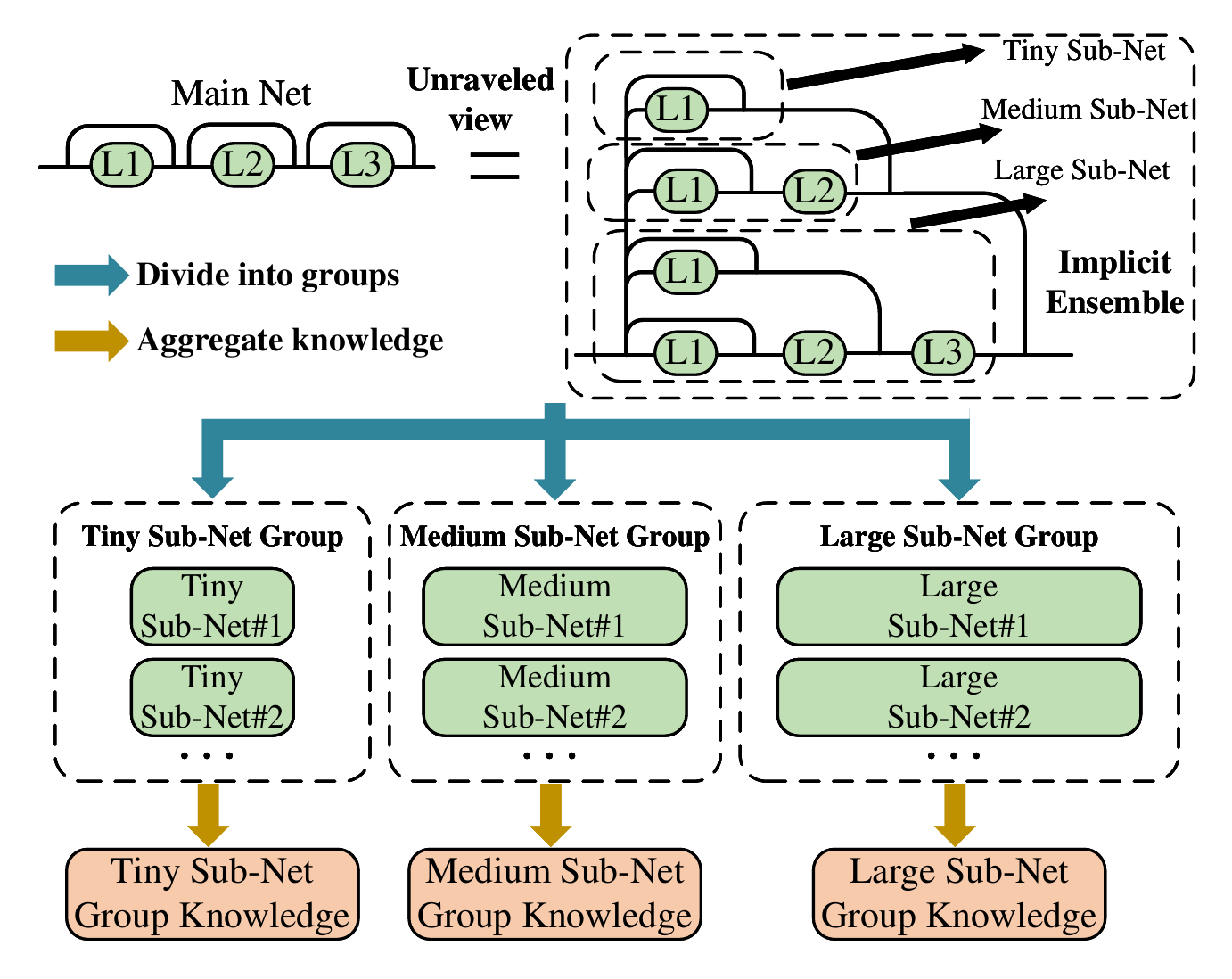}
  \caption{Illustration of \textbf{unraveled view} and \textbf{group knowledge}. Unraveled view shows that residual networks can be seen as an ensemble of numerous networks of different lengths. Inspired by this viewpoint, we allocate the subnets into subnet groups of different sizes, i.e. tiny, medium, and large subnet groups in the figure. Then we exploit the knowledge of subnet groups during training to boost the performance of given residual networks effectively and efficiently.}

  \label{fig:unraveled view}
\end{figure}

Residual structures, first introduced in~\cite{he2016deep}, have become nearly indispensable in mainstream network architectures. It achieved great success in numerous architectures, such as convolutional networks~\cite{tan2021efficientnetv2, ye2022beta, ye2022efficient, ye2023beta, mei2023automatic}, recurrent networks~\cite{galshetwar2022lrnet}, MLP networks~\cite{tolstikhin2021mlp}, and transformers~\cite{vaswani2017attention, huang2023experts, liang2023rethinking}. Considering the extraordinary performance of the residual structure, it is drawing increasing attention~\cite{he2020resnet, ding2022overparameterization, barzilai2022kernel} to study the underlying mechanisms leading to their success. An interesting explanation is that residual networks can be regarded as an implicit ensemble of relatively shallow subnets, namely \textbf{unraveled view}~\cite{veit2016residual, barzilai2022kernel}. It opens up a new perspective to further improve the performance of residual networks. One of the common ways is to randomly sample subnets and train them individually. Stochastic depth ~\cite{huang2016deep} randomly drops a subset of layers and trains the remaining layers with ground truth labels. \cite{yestimulative} have observed a phenomenon known as ``network loafing", where the standard training procedure fails to provide sufficient supervision and often causes subpar performance. To address the above problem,~\cite{yestimulative} propose a stimulative training(ST) strategy by training randomly sampled subnets with outputs from the main network. However, these methods train various subnets with \yep{the same kind of supervision (i.e., stochastic depth always uses the ground-truth, and ST always uses the output of the main network)}, regardless of their unique capacities. It is straightforward to investigate \textbf{whether applying \yep{the same kind of supervision} for diverse subnets is suitable}.
Inspired by~\cite{mirzadeh2020improved}, we believe that ``suitable supervision"
should meet two important criteria: (1) \textbf{easy to be transferred (with a limited capacity gap),} (2) \textbf {containing rich and useful knowledge}. 

To reduce the capacity gap, a common method~\cite{mirzadeh2020improved} is to introduce extra intermediate models as teacher assistants. To get abundant knowledge, self knowledge distillation can be applied to learn from prior experience/knowledge, and ensemble knowledge distillation may further combine the supervisions of various teachers. Under the novel unraveled view, there are numerous subnets with various capacities, and the grouped assistant teachers naturally exist. Inspired by the observations above, we divide all subnets of a residual network into multiple groups by their capacity and aggregate their abundant knowledge during training, as shown in Figure~\ref{fig:unraveled view}.
\tsj{Interestingly, in the sociology field, transferring suitable knowledge is also important for improving the productivity of organizations~\cite{baum1998survival}. 
Group knowledge~\cite{kane2005knowledge}, collected from the same producing group, is considered easier to transfer to members in the neighboring group. Similar to the group knowledge in the field of sociology, we aggregate the knowledge from different subnets in the same group to build suitable supervision, and vividly call the aggregated knowledge as \textbf{network group knowledge}. Generally speaking, the knowledge produced by multiple subnet groups has two excellent properties:
(1) it is naturally hierarchical and easy to be utilized to fill the capacity gap; (2) it is aggregated by numerous subnets containing abundant knowledge.} 
\par
\begin{figure}[t]
  \centering
  \includegraphics[width=\linewidth]{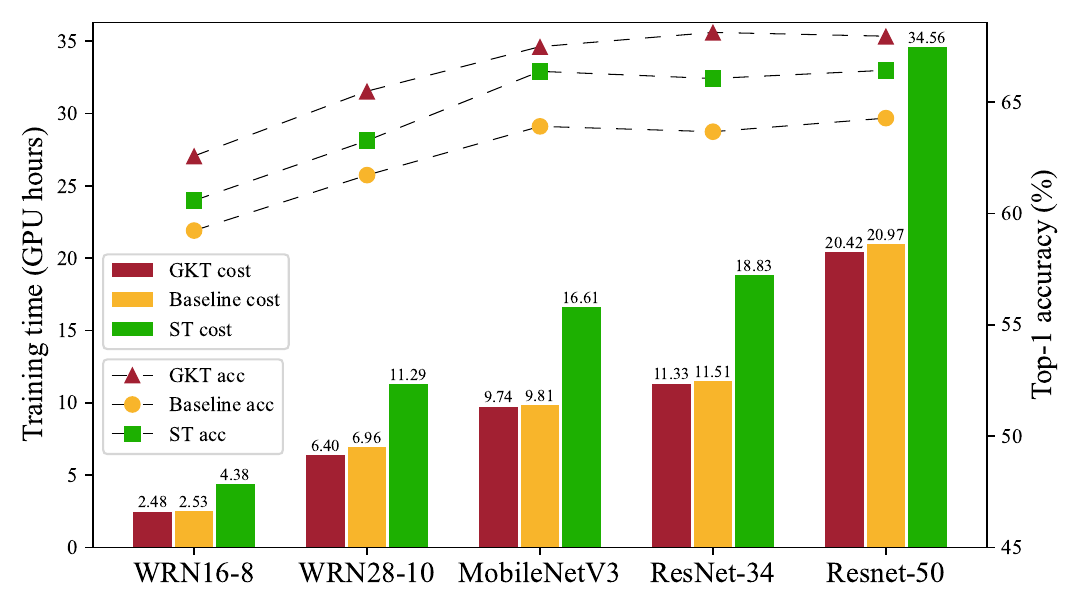}
  \caption{Training time and accuracy on TinyImageNet for group knowledge based training strategy, and other methods like standard training and stimulative training (ST).}
  \label{fig:effectiveness and efficiency}
\end{figure}

 
Based on the findings above, we further propose the \textbf{group knowledge based training (GKT)} framework, \yep{for boosting the performance of residual networks effectively and efficiently. In detail, during the training procedure, we first divide all subnets of a residual network into hierarchical subnet groups by a sampling strategy called \lwh{subnet-in-subnet (SIS)} sampling, then aggregate the knowledge of subnets in the same group by network logits moving average, and then supervise the subnet with an appropriate level of group knowledge. Moreover, we find that sampling and training larger subnets can better boost the performance of residual networks, thus we design an inheriting exponential decay rule}
to focus on \bp{the  large  or medium subnets}. 
\yep{The proposed GKT framework} can remarkably boost the network performance without any extra parameter (e.g., assistant teacher) or heavy computation cost (e.g., \lwh{forwarding} main net to obtain supervision). \yep{The efficacy and efficiency of GKT is shown in Figure~\ref{fig:effectiveness and efficiency}.} GKT does not require model topological modifications and only samples a part of a network (subnet) in the training procedure, resulting in less inference cost and training time compared with standard training (i.e., shown in the baseline of Figure \ref{fig:effectiveness and efficiency}). \tsjaddd{Because most CNN and transformer models adopt residual architecture and suffer from network loafing~\cite{ye2023stimulative}, we further verify GKT on various CNN and transformer models.}

\par In summary, our contributions are as follows:
\begin{itemize}
\item \yep{From the novel unraveled view of the residual network, we identify the hierarchical subnet group knowledge for the first time, which can provide better supervision for the diverse subnets of the residual network.}
\item \yep{We propose the GKT framework for boosting residual networks effectively and efficiently. In this framework, subnet-in-subnet sampling is adopted to implicitly divide all subnets into hierarchical subnet groups. Subnet logits' exponential moving average is exploited to aggregate the knowledge in the same subnet group.}
\item 
 \yep{We experimentally verify sampling and training larger subnets can benefit residual networks more than smaller subnets. Thus we design an inheriting exponential decay rule to sample larger subnets for training subnets and preparing subnet groups.} 
\item \tsjaddd{Comprehensive empirical comparisons and analysis show that GKT can reduce the capacity gap and efficiently improve the performance of various residual networks, including CNNs and transformers.}  
\end{itemize}

\section{Related Works}
\subsection{Unraveled View}
\label{sec:unraveled view related work}
\par To better understand residual networks,~\cite{veit2016residual} introduces a novel perspective named unraveled view, that interprets residual networks as \yep{an ensemble} of shallower networks (i.e., subnets).
Based on unraveled view, ~\cite{sun2022low} verifies that \yep{shallow and deep subnets correspond to the low-degree and high-degree polynomials \lwh{respectively}, and shallow subnets play important roles when training residual networks. Then,~\cite{barzilai2022kernel} theoretically proves that the eigenvalues of the residual convolutional neural tangent kernel (CNTK) are made of weighted sums of eigenvalues of CNTK of subnets.} Based on the unraveled view,~\cite{huang2016deep} directly trains random subnets to improve the performance of residual networks.~\cite{yestimulative} reveals the network loafing problem that standard training causes serious subnet performance degradation, and proposes stimulative training (ST) to supervise all subnets by the main net.
\par Following the research stream of unraveled view, we \yep{also} focus our investigation on boosting residual networks by improving their subnets.
Different from providing the same kind of supervision (e.g., ground truth~\cite{huang2016deep} or main net logits~\cite{yestimulative}) for all subnets, we pay attention to giving different subnets different supervisions according to their model capacities. \yep{Besides, we discover that the training of relatively large subnets can benefit the main net more, and we focus more on supervising the large or medium subnets instead of all subnets.}

\subsection{Knowledge Distillation}
\label{sec:kd}
\yep{Conventional Knowledge distillation (KD)~\cite{hinton2015distilling} transfers knowledge from a teacher network to a student network via logits~\cite{kim2018paraphrasing, shen2022fast} or features~\cite{bai2020feature,jung2021fair}, 
aiming at obtaining a compact and accurate student network. It usually requires additional cost because of training a larger teacher network. As a comparison, we do not require larger teachers or additional structures. And our target is to improve any given residual network effectively and efficiently by training its subnets well. Most related concepts among KD are described in detail as follows.}
\par \noindent\textbf{Ensemble Distillation.} As the ensemble method~\cite{dietterich2000ensemble} is a useful technique to improve the performance of deep learning models, it is generally considered that, an ensemble of multiple teacher models can commonly provide supervision with higher quality compared with a single teacher.
~\cite{du2020agree} studies the conflicts and competitions among teachers and introduces \yep{a dynamic weighting method for better fusing teachers' knowledge.}
However, typical ensemble distillation methods need additional teacher models to provide supervision for a single student model. It requires huge computation cost and is not suitable for
multiple coupled subnet students.
Differently, we do not need extra teacher models or inference. For multiple unique coupled subnet students, we specifically provide suitable supervision by aggregating the hierarchical subnet group knowledge. 

\par \noindent\textbf{Self Distillation.} To save the cost introduced by a larger teacher network, the self distillation (SD) attempts to provide supervision within the student network itself in training.
\cite{yun2020regularizing} \yep{narrows down} the predictive distribution deviation between different samples of the same label to provide the regularization supervision.~\cite{deng2021learning, kim2021self} utilize the snapshot of the previous output logits as supervision to learn from prior experience.~\cite{shen2022self} \yep{rearranges the data sampling by including mini-batch from previous iteration, 
and uses the on-the-fly soft targets generated in the previous iteration to supervise the network.} Similarly, we also consider that the historic information during the training contains abundant knowledge. However, previous SD methods only utilize the intermediate features or output logits of a single main net. Differently, motivated by the unraveled view, we focus on various subnets with distinct capacities and utilize their aggregated knowledge in different iterations.

\par \noindent \textbf{\tsjadd{Online Distillation.}}
\tsjadd{Online knowledge distillation (OKD) introduces extra multiple branches or models manually during the training procedure to extract knowledge. ONE~\cite{zhu2018knowledge} introduces additional branches to create a native ensemble teacher and transfer the knowledge from the ensemble teacher to each branch. PCL~\cite{wu2021peer} builds ensemble teachers by integrating different branches and meaning them temporally to supervise each branch. OKDDip~\cite{chen2020online} proposes to enhance the diversity of multiple branches with attention-based weights. Different from OKD\cite{wu2021peer, chung2020feature} using the same knowledge-transfer strategies for each fixed student, we focus on providing tailored knowledge for dynamically sampled subnets with a lower capacity gap. Moreover, GKT aggregates intrinsic knowledge \textbf{without any extra architecture} causing easier implementation for different architectures and less training cost.}

\par \noindent\textbf{Capacity Gap.} There is a counter-intuitive phenomenon ~\cite{cho2019efficacy} called capacity gap, referring to the fact that a larger and more accurate teacher model does not necessarily teach \yep{the} student model better.
This phenomenon is attributed to the capacity mismatch, that a tiny student model has insufficient ability to mimic the behavior of a large teacher with huge capacity. For transferring knowledge better, numerous works are proposed to bridge the capacity gap.\yep{~\cite{mirzadeh2020improved} introduces extra intermediate models as teacher assistants.~\cite{li2022asymmetric, guo2022reducing} propose asymmetric temperature scaling for teacher and student to make larger teachers teach better}. Since there are subnets with different capacities under the unraveled view, the capacity gap problem becomes more serious when transferring knowledge to various subnets. Different from the above methods, we bridge the capacity gap by aggregating the hierarchical subnet group knowledge, without additional models~\cite{mirzadeh2020improved} or changes of hyper-parameters like temperature~\cite{li2022asymmetric, guo2022reducing}.
\begin{figure*}[]
  \centering
  \includegraphics[width=0.98\linewidth]{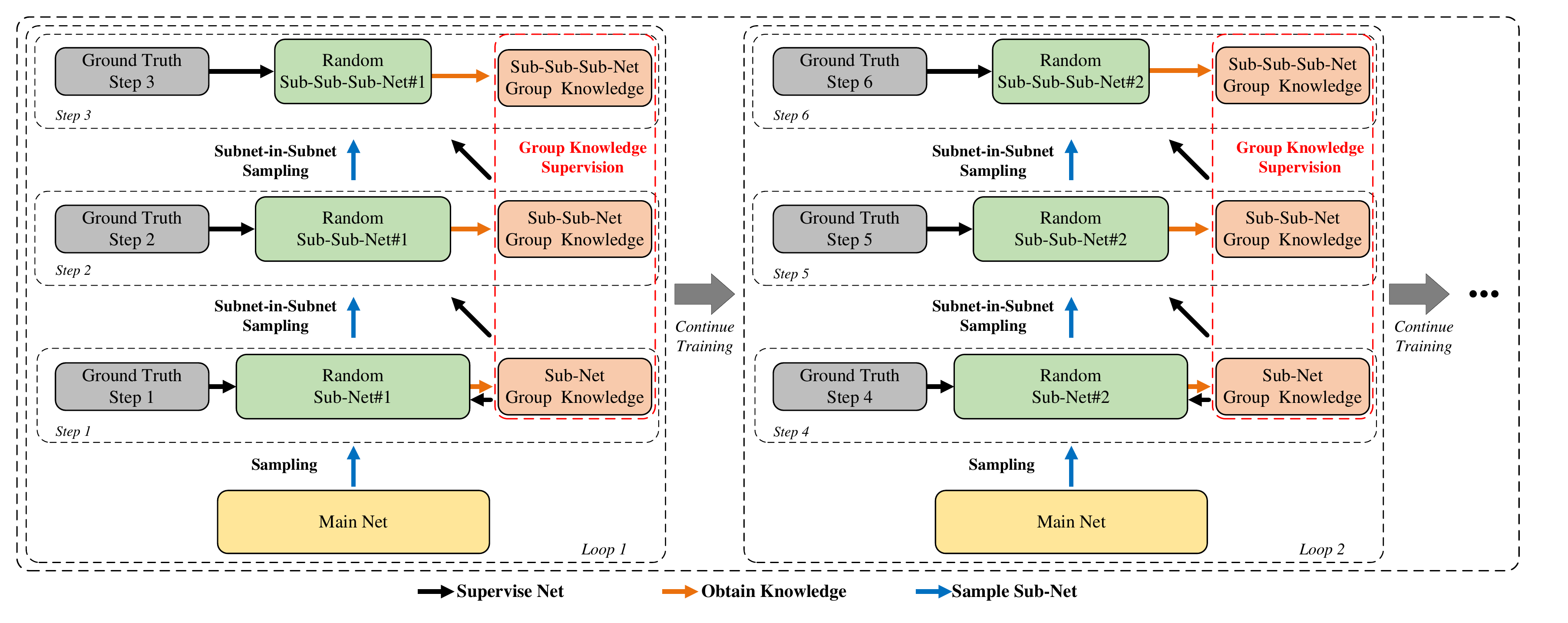}
  \caption{Overview of the group knowledge based training (GKT) framework. To give an example, we suppose there are three groups. GKT divides all subnets into hierarchical subnet groups by subnet-in-subnet sampling, aggregates the knowledge of different subnets of different steps in the same subnet group, and uses the aggregated group knowledge to supervise the neighboring subnet group. To avoid continuously sampling the tiniest subnet, multiple sampling loops are applied to successively sample subnets from the main network again, as expressed in the left part (i.e., loop 1) and the right part (i.e., the next loop 2). More details of subnet-in-subnet sampling are shown in \textbf{Appendix} Fig. r5.
  }
  \label{fig:compare_ideas}
\end{figure*}

\section{Group Knowledge based Training (GKT)}

\subsection{Framework}
\par The overview of group knowledge based training (GKT) is shown in Figure~\ref{fig:compare_ideas}. We divide the total number of training steps by the given number of subnet groups to get multiple equal training loops. 
For each loop, the operations of GKT consist of three parts: (1) Subnet Group Division; (2) Group Knowledge Aggregation; (3) Group Knowledge Transfer.
\par \emph{Subnet Group Division}: At the beginning of each loop, we sample a subnet from the main net. Then, we continue to sample a subnet from the parent subnet, until the end of the loop. All subnets sampled in the same generation on different loops are divided into the same group. After sampling, we forward the subnet to obtain logits and compute the loss.
\par \emph{Group Knowledge Aggregation:} The subnet logits, generated on the specific step of each loop, are used to update the subset's corresponding group knowledge by logit-level exponential moving average.

\par \emph{Group Knowledge Transfer:}
At each training step, the ground truth and neighboring larger group knowledge will be utilized to supervise the current subnet. \tsj{The pseudo-code of GKT is shown in \textbf{Appendix B1}.}
\par During testing, GKT forwards the residual network without modifying the structure or changing the testing pipeline. 
We will introduce these three parts in detail as follows.


\par \noindent\textbf{Group Division: Subnet-in-Subnet Sampling.}
To alleviate the capacity gap when supervising diverse subnets, we propose to divide subnets into hierarchical groups. An intuitive division method is \yep{to randomly sample subnets and directly divide them by their parameters or FLOPs.} However, these direct methods introduce \bp{many} hyper-parameters, such as partition bounds of each group, and  \bp{may} limit the number of possible subnets in a group. Therefore, we introduce \yep{subnet-in-subnet (SIS) sampling} 
to naturally divide all subnets into hierarchical subnet groups. \yep{Specifically, SIS sampling strategy} samples subnet from the parent subnet and divides the subnet sampled in the same generation into the same group. \yep{
Besides, to avoid being restricted to tiny subnets due to an unending sampling, the total training steps are equally partitioned into several loops, and at the beginning of each loop, the sampling process starts from the main net.} 
\par 
\yep{Formally, we denote the subnet $\mathcal{N}_s$ belonging to the $t$-th group and sampled in the $r$-th loop as $\mathcal{N}_{s,t}^r$}. Given a parent subnet $\mathcal{N}_{s,t}^r$, the next sampled subnet is generated as 
\begin{align}
\label{formulation:SIS}
\mathcal{N}_{s,t+1}^r = \pi(\mathcal{N}_{s,t}^r),
\end{align}
where $\pi(.)$ is the sampling operation representing randomly sampling a subnet from a network \yep{based on} given sampling distribution. At the beginning of a loop, \yep{we sample a subnet from the main net}, expressed formally as $\mathcal{N}_{s,1}^r = \pi(\mathcal{N}_{m})$. \yep{Then, we continue to sample one subnet from the parent subnet at each step, as shown in Equation~\ref{formulation:SIS}.} 
The subnets sampled in the \yep{$t$-th} generation \ct{step} in all loops are regarded in the \yep{$t$-th} group. After $M$ sampling \ct{steps}, the current training loop ends\yep{, and the sampling} comes to the next training loop. It is noticed that a specific subnet might be sampled in any loop and \bp{divided} into any groups. \yep{Subnet-in-subnet sampling} utilizes the inheritance relationship of subnets as a criterion for loose group division, \yep{and its surpassing effectiveness is verified in \textbf{Appendix D2}}.

\begin{figure}[t]
  \centering
  \includegraphics[width=0.96\linewidth]{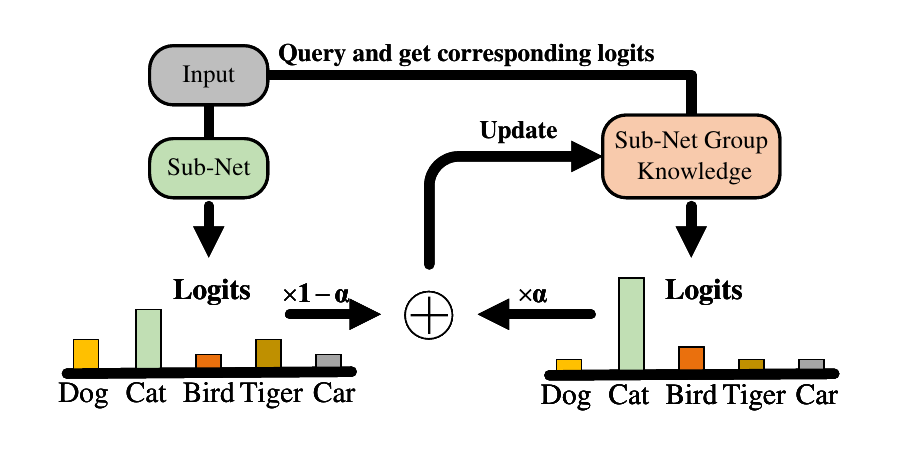}
  \caption{Updating mechanism of subnet group knowledge. We use the exponential moving average (EMA) of network logits to update the subnet group knowledge.
  }
  \label{fig:EMA}
\end{figure}

\par \noindent\textbf{Group Knowledge Aggregation: Subnet Logits Exponential Moving Average.} \yep{As network logits are the most commonly used supervision containing substantial high-level knowledge, we save and aggregate output logits of different subnets in the same group} as group knowledge. However, \yep{considering that the subnets in the same group distribute in different temporal frames,} it is unbearable to save all of their historic logits. Inspired by model parameter exponential moving average (EMA) in self-supervised learning~\cite{he2020momentum}, we adopt subnet logits EMA to aggregate group knowledge effectively and resource-friendly. 
\par Formally, we denote $\mathcal{K}_t \in \mathbb{R}^{N \times k}$, where $N$ and $k$ are the data and class number of \yep{the dataset}, as the \yep{$t$-th} group knowledge. As shown in Figure~\ref{fig:EMA}, \yep{for mini-batch samples $x \in \mathbb{R}^{b \times c \times h \times w}$, we denote $\mathcal{K}_t^I \in \mathbb{R}^{b \times k}$ as the corresponding group knowledge queried from $\mathcal{K}_t$ by indices $I$. Then the} corresponding subnet group knowledge is updated as 
\begin{align}
\label{formulation:EMA}
\mathcal{K}_t^I := \alpha p(\theta_{\mathcal{N}_{s,t}},x) + (1-\alpha) \mathcal{K}_t^I
\end{align}
where $:=$ represents updating and $\alpha$ is the EMA coefficient to balance the intensity of updating. If it is \bp{the} first time to update $\mathcal{K}_t^I$, we directly initialize it with $p(\theta_{\mathcal{N}_{s,t}},x)$. It is worth noting that because we can store $\mathcal{K}$ into the disk instead of GPU memory, subnet logits EMA only introduces negligible storage and computation cost.

\begin{table*}[h]
\centering
\begin{tabular}{l|lccccc}
\hline
Dataset                       & Method   & WRN16-8        & WRN28-10       & MobileNetV3    & ResNet-34      & ResNet-50      \\ \hline
\multirow{8}{*}{CIFAR-100}    & Baseline       & 79.95          & 82.17          & 78.09          & 77.78          & 78.14          \\
                              & StoDepth~\cite{huang2016deep} & 80.64          & 82.75          & 78.77          & {\underline {80.62}}    & 78.43          \\
                              & ONE~\cite{zhu2018knowledge}                        & 80.49               & \underline{83.02}               & 80.85                & 79.96     
                                     & 80.44                \\
                              & CS-KD~\cite{yun2020regularizing}   & 80.77          & 81.25          & 78.60           & 79.35          & 80.34          \\

                              & PS-KD~\cite{kim2021self}   & {\underline{81.17}}    & 82.53          & 79.36          & 79.30          & 80.07          \\
                              & LWR~\cite{deng2021learning}     & 81.05          & 82.00          & 80.23          & 79.81          & 80.70           \\
                              & DLB~\cite{shen2022self}     & 80.87          & 81.35          & 78.55          & 79.26          & 80.36          \\
                              & ST~\cite{yestimulative}      & 80.75          & {82.84}    & {\underline {81.07}}    & 78.62          & {\underline {81.06}}    \\ \cline{2-7} 
                              & GKT     & \textbf{81.53} & \textbf{84.38} & \textbf{81.70}  & \textbf{81.40}  & \textbf{81.73} \\ \hline
\multirow{8}{*}{TinyImageNet} & Baseline       & 59.23          & 61.72          & 63.91          & 63.67          & 64.28          \\
                              & StoDepth~\cite{huang2016deep} & 60.29          & 62.02          & 64.97          & 65.75          & 65.80           \\
                              & ONE~\cite{zhu2018knowledge}   & \underline{62.13}                & 64.15               & 65.39                & \underline{66.98}     
                                     & 66.58                \\
                              & CS-KD~\cite{yun2020regularizing}   & 60.23          & 62.24          & 64.72          & {66.11}    & {\underline {66.74}}    \\
                              & PS-KD~\cite{kim2021self}   & 60.93          & 63.23          & 66.41          & 65.77          & 65.96          \\
                              & LWR~\cite{deng2021learning}     & {61.62}    & 63.91          & {\underline {66.43}}    & 63.51          & 65.03        \\
                              & DLB~\cite{shen2022self}     & 61.48          & {\underline {64.29}}    & 65.05          & 65.86          & 65.78          \\
                              & ST~\cite{yestimulative}      & 60.58          & 63.27          & 66.38          & 66.06          & 66.43          \\ \cline{2-7} 
                              & GKT     & \textbf{62.58} & \textbf{65.49} & \textbf{67.49} & \textbf{68.13} & \textbf{67.96} \\ \hline
\end{tabular}
\caption{Main experimental results of the proposed GKT and other methods on the \textbf{CIFAR-100} and \textbf{TinyImageNet} datasets. The best performance is highlighted in bold, and the second-best performance is highlighted in underline.}
\label{table:main results}
\end{table*}

\begin{table*}[t]
\centering
\begin{tabular}{l|cccc|cccccc}
\hline
\multirow{2}{*}{Method} & \multicolumn{2}{c}{Resnet-34}                                                                                    & \multicolumn{2}{c|}{Resnet-50}                                                                                   & \multicolumn{2}{c}{Swin-T}                                                                                       & \multicolumn{2}{c}{Swin-S}                                                                                       & \multicolumn{2}{c}{ViT-S}                                                                                        \\ \cline{2-11} 
                        & \begin{tabular}[c]{@{}c@{}}Top-1\\ Acc(\%)\end{tabular} & \begin{tabular}[c]{@{}c@{}}Cost\\ (hours)\end{tabular} & \begin{tabular}[c]{@{}c@{}}Top-1\\ Acc(\%)\end{tabular} & \begin{tabular}[c]{@{}c@{}}Cost\\ (hours)\end{tabular} & \begin{tabular}[c]{@{}c@{}}Top-1\\ Acc(\%)\end{tabular} & \begin{tabular}[c]{@{}c@{}}Cost\\ (hours)\end{tabular} & \begin{tabular}[c]{@{}c@{}}Top-1\\ Acc(\%)\end{tabular} & \begin{tabular}[c]{@{}c@{}}Cost\\ (hours)\end{tabular} & \begin{tabular}[c]{@{}c@{}}Top-1\\ Acc(\%)\end{tabular} & \begin{tabular}[c]{@{}c@{}}Cost\\ (hours)\end{tabular} \\ \hline
Baseline                & 74.70                                                   & 98.04                                                  & 76.98                                                   & 202.16                                                 & 77.50                                                   & 301.47                                                 & 79.36                                                   & 460.82                                                 & 75.55                                                   & 301.91                                                 \\
StoDepth                & 74.96                                                   & \textbf{94.76}                                         & 77.43                                                   & 196.77                                                 & 79.62                                                   &     297.74                                             & 81.23                                                   & 452.48                                                & 77.03                                                   & 296.41                                                 \\
ST                      & 75.25                                                   & 166.98                                                 & 77.60                                                   & 345.45                                                 & 79.87                                                   & 437.23                                                 &  81.43                                                        &   685.28                                                     & 77.27                                                   & 447.77                                                 \\ \hline
GKT                     & 75.50                                                   & 95.48                                                  & 78.10                                                   & \textbf{194.32}                                        & 80.40                                                   & \textbf{294.38}                                        & 82.00                                                   & \textbf{451.18}                                        & 78.51                                                   & \textbf{295.62}                                        \\
GTK (+epoch)            & \textbf{75.83}                                          & 141.13                                                 & \textbf{78.40}                                          & 290.36                                                 & \textbf{80.72}                                                        &   442.55                                                     &  \textbf{82.26}                                                       & 678.24                                                       & \textbf{78.83}                                                        & 445.78                                                       \\ \hline
\end{tabular}
\caption{Verification of CNNs and transformers on the \textbf{ImageNet-1K} dataset.}
\label{table:img_1k}
\end{table*}

\begin{table*}[htbp]
\begin{center}
\begin{tabular}{c|ccccccc}
\hline
         & ViT\_C10      & Swin\_C10      & CaiT\_C10      & ViT\_C100      & Swin\_C100     & CaiT\_C100     & ViT\_TinyImg   \\ \hline
Baseline & 93.23         & 94.05          & 94.71          & 72.15          & 76.02          & 76.52          & 54.14          \\
StoDepth & 93.58         & 94.46          & 94.91          & 73.81          & 76.87          & 76.89          & 57.07          \\ \hline
GKT      & \textbf{93.8} & \textbf{95.21} & \textbf{95.24} & \textbf{75.31} & \textbf{77.32} & \textbf{78.79} & \textbf{58.71} \\ \hline
\end{tabular}
\caption{\centering Verification of various \textbf{transformers} on \textbf{CIFAR-10/100(C10/100)} and \textbf{Tiny ImageNet(TinyImg)}.}
\label{table:cifar transformer}
\end{center}
\end{table*}

\begin{table*}[ht]
{
\begin{tabular}{c|c|c|cc|ccc}
\hline
\multirow{2}{*}{} & ResNet-50      & Faster R-CNN(Det) & \multicolumn{2}{c|}{Mask R-CNN(Det\&Seg)} & \multicolumn{3}{c}{Panoptic FPN(Panotic Seg)}    \\ \cline{2-8} 
                  & ImageNet Acc   & Det mAP@0.5       & Det mAP@0.5         & Seg mAP@0.5         & PQ             & SQ             & RQ             \\ \hline
Baseline   & 76.98          & 59.4              & 59.6                & 56.5                & 41.76          & 78.21          & 51.38          \\
GKT    & \textbf{78.10} & \textbf{60.1}     & \textbf{60.1}       & \textbf{57.3}       & \textbf{42.27} & \textbf{78.83} & \textbf{51.62} \\ \hline
\end{tabular}
}
\caption{Verification on object detection, instance segmentation and panoptic segmentation with schedule 1× on COCO2017.}
\label{table:downstream tasks}
\end{table*}

\par \noindent\textbf{Hierarchical Group Knowledge Transfer.}
After group knowledge aggregation, there is a group knowledge pool containing \yep{different levels of} group knowledge. 
To reduce the capacity gap and obtain abundant knowledge, we transfer the neighboring larger group knowledge to the subnet. Formally, for a given subnet $\mathcal{N}_{s,t}$ in \yep{$t$-th} group, the supervision is $\mathcal{K}_{t-1}^I$, \yep{the loss is Kullback-Leible divergence} 
\begin{align}
\label{formulation:transfer}
\mathcal{L}_{GK} = KL(\mathcal{K}_{t-1}^I,p( \theta_{\mathcal{N}_{s,t}} ,x)).
\end{align}
And the total loss of GKT is the weighted sum of group knowledge supervision and standard cross-entropy loss 
\begin{small}
\begin{align}
\label{formulation:GKT loss}
\mathcal{L}_{GKT} = CE(p(\theta_{\mathcal{N}_{s,t}},x ),y) + \beta KL(\mathcal{K}_{t-1}^I,p( \theta_{\mathcal{N}_{s,t}} ,x)).
\end{align}
\end{small}
$\beta$ is a loss balanced coefficient. For subnets in the largest group, they are supervised by their own group knowledge.

\par \noindent\textbf{Inheriting Exponential Decay Rule.}
Under the unraveled view, there are $2^L$ subnets \yep{in the given residual network}, where $L$ is the number of residual blocks, which is a huge sampling space. It's almost impossible to train every subnet sufficiently. 
\yep{Thus,}~\cite{yestimulative} proposes to keep the ordered residual structure of subnets for sampling space reduction and easier subnet training.~\cite{huang2016deep} follows the intuition that the earlier blocks extract more important low-level features and are more reliably present, 
\yep{thus adopting} linear decay sampling to drop the deeper blocks with higher probability. \yep{Furthermore, we experimentally verify} that \yep{training} relatively large subnet benefits \yep{the} main net more, which will be discussed \yep{detailedly} in Section~\ref{sec:train large}. 
\par \yep{Inspired by these,  we propose an inheriting exponential decay subnet sampling strategy. To ensure larger subnets are sampled with a higher probability, we propose to change the sampling distribution of global block-wise linear decay~\cite{huang2016deep} to stage-wise exponential decay.}
\yep{Since we utilize the} subnet-in-subnet sampling, \yep{our} sampling distribution is dynamic \yep{w.r.t. the} parent subnet, \yep{thus} called ``inheriting". \yep{Besides, we also keep the ordered residual structure when sampling.} The effectiveness of inheriting exponential decay sampling is verified in \textbf{Appendix D4}. 
\par Formally, for a parent subnet $\mathcal{N}_{s,t}$ belonging to \yep{$t$-th} group, it is usually made up of several stages, and each stage contains several residual blocks. We suppose $\mathcal{N}_{s,t}$ has $D$ stages and  the block number of \yep{each} stage is $[l_1, l_2, ... l_D]$.
For \yep{each stage,} we utilize a sampling distribution corresponding to the number of retained ordered residual blocks \yep{to control the sampling process}. 
To reduce hyper-parameters, we give a total base number $q \in (0,1)$, and the sampling distribution of \yep{$d$-th} stage, i.e. $\chi_d$, corresponding to block number $[1,2, ... l_d]$, is computed as 
\begin{align}
    \label{formulation:exponential1}
    &u = q^{D-d+1}\\
    \label{formulation:exponential2}
    &v = [u^{l_d}, u^{l_d-1}, ... u] \\
    \label{formulation:exponential3}
    &\chi_d = [\frac {u^{l_d}} {\sum v}, \frac {u^{l_d-1}} {\sum v}, ... \frac {u} {\sum v}]
\end{align}
where $u$, $v$ are temporary \yep{values. Superscript} represents the exponent. From \yep{Equation}~\ref{formulation:exponential2} and ~\ref{formulation:exponential3}, we can observe \yep{that} the probability of sampling \yep{larger subnets from any parent network} has been remarkably increased. 

 \begin{table}[ht]
\centering
\begin{tabular}{ccc|c}
\hline
SIS Sampling & SL-EMA & HGKT & Top-1 Acc(\%) \\ \hline
\checkmark   & \checkmark               & \checkmark             & \textbf{84.38}         \\
\checkmark   & \checkmark               & $\times$              & 83.51         \\
\checkmark   & $\times$                & $\times$             & 82.40          \\
$\times$    & $\times$                & $\times$              & 82.17         \\ \hline
\end{tabular}
 \caption{Influence of different components including Subnet-in-Subnet (SIS) Sampling, Subnet Logits EMA (SL-EMA), Hierarchical Group Knowledge Transfer (HGKT).}
 \label{table:ablation}
\end{table}

\begin{table}[ht]
\centering
\begin{tabular}{l|l|l}
\hline
Sampling Strategy & ST             & GKT            \\ \hline
UR(1$\sim$12)     & 82.84          & 81.06          \\
UR(5$\sim$6)      & 81.51          & 80.58          \\
UR(7$\sim$8)      & \textbf{82.91} & 82.32          \\
UR(9$\sim$10)     & 82.76          & 82.93          \\
UR(11$\sim$12)    & 82.58          & 83.51          \\ \hline
EDR(1$\sim$12)    & 82.9           & \textbf{84.38} \\ \hline
\end{tabular}
\caption{\tsj{Influence of different sampling strategies on the final performance. To explore the influence of sampling space, we implement ST and GKT with different sampling rules and spaces, including uniform rule (UR) on several different spaces and Exponential Decay Rule (EDR) on the full space.}}
\label{table:mini-spaces}
\end{table}

\section{Experiments}
\par \tsjadd{We first verify the effectiveness and efficiency of GKT on image classification with CNNs and transformers. To further demonstrate the generality of GKT, we conduct experiments on downstream tasks. Then, ablation experiments show the indispensability of each component. Finally, investigation experiments reveal the mechanism of GKT. The details of experiment settings are shown in \textbf{Appendix A}.}

\subsection{Image Classification}
We demonstrate the effectiveness and efficiency 
of GKT on typical residual convolutional networks including ResNet-34, ResNet-50~\cite{he2016deep}, WRN16-8, WRN28-10~\cite{zagoruyko2016wide} and MobileNetV3~\cite{howard2019searching}, and mainstream datasets including CIFAR-100, Tiny ImageNet and ImageNet-1K. Observing that residual connections popularly exist in visual transformers, we conduct experiments on transformers including ViT~\cite{dosovitskiyimage}, Swin~\cite{liu2021swin} and CaiT~\cite{touvron2021going} to prove the generalization ability further. To verify the superiority of GKT, we compare the test accuracy with standard training, subnet training methods, i.e., ST~\cite{yestimulative} and Stodepth~\cite{huang2016deep}, prevailing SD methods, i.e., CS-KD~\cite{yun2020regularizing}, PS-KD~\cite{kim2021self}, DLB~\cite{shen2022self} and LWR~\cite{deng2021learning}, and online distillation method. i.e., ONE~\cite{zhu2018knowledge}. 
\par The results on \textbf{CIFAR-10/100} and \textbf{Tiny ImageNet} are shown in Table~\ref{table:main results} (on various CNNs) and Table~\ref{table:cifar transformer} (on various transformers). For \textbf{CNNs}, GKT universally and remarkably boosts the performance of different residual networks on different datasets. To be more specific, compared with standard training, the average Top-1 test accuracy improvements of GKT on different networks are up to 1.67\% on CIFAR-100 and 1.22\% on Tiny ImageNet, respectively. Besides, compared with other SD and subnet training methods, GKT consistently achieves a new state-of-the-art performance, which demonstrates the superiority of GKT over other approaches. Specifically, the Top-1 test accuracy gains of GKT compared with the second-best method are up to 1.54\% on CIFAR-100 and 2.02\% on Tiny ImageNet respectively. For \textbf{transformers}, it is observed that GKT remarkably improves the accuracy e.g., +4.57\% on Tiny ImageNet. This verifies the potential of GKT to boost different residual architectures. 
\par We also verify the \yep{generalization ability} and efficiency of GKT on \yep{\textbf{ImageNet-1K}}, i.e. a mainstream large scale dataset. As shown in Table~\ref{table:img_1k}, both in CNNs and transformers, GKT can achieve significant performance gains over the baseline of different networks, and perform better than Stodepth and ST. Meanwhile, the time cost of GKT is almost the smallest among these methods. Besides, the performance of different networks can be further boosted when increasing the training epoch of GKT.

\begin{table}[t]
\centering
\begin{tabular}{l|ccc|c}
\hline
Group Rank & Tiny & Medium & Large & Expectation \\ \hline
Group\#1   & 0.35             & 0.34               & 0.31               & 21.54M                 \\
Group\#2   & 0.19             & 0.29               & 0.52               & 25.50M                 \\
Group\#3   & 0.06             & 0.15               & 0.79               & 30.43M                 \\ \hline
\end{tabular}
\caption{Properties of different subnet groups obtained by subnet-in-subnet sampling \tsj{on WRN28-10 (36.54M)}. We show the sampling ratio of tiny (7.4$\sim$17.17M), medium (17.17$\sim$26.85M), and large(26.85$\sim$36.54M) subnets for different subnet groups and give the parameter expectation.}
\label{table:subnet parameters}
\end{table}

\subsection{Ablation Experiments}
\label{sec:ablation experiments}

\begin{figure}[t]
\centering  
\subfigure[ResNet-50]{
\label{Fig.sub.2}
\includegraphics[width=0.47\linewidth]{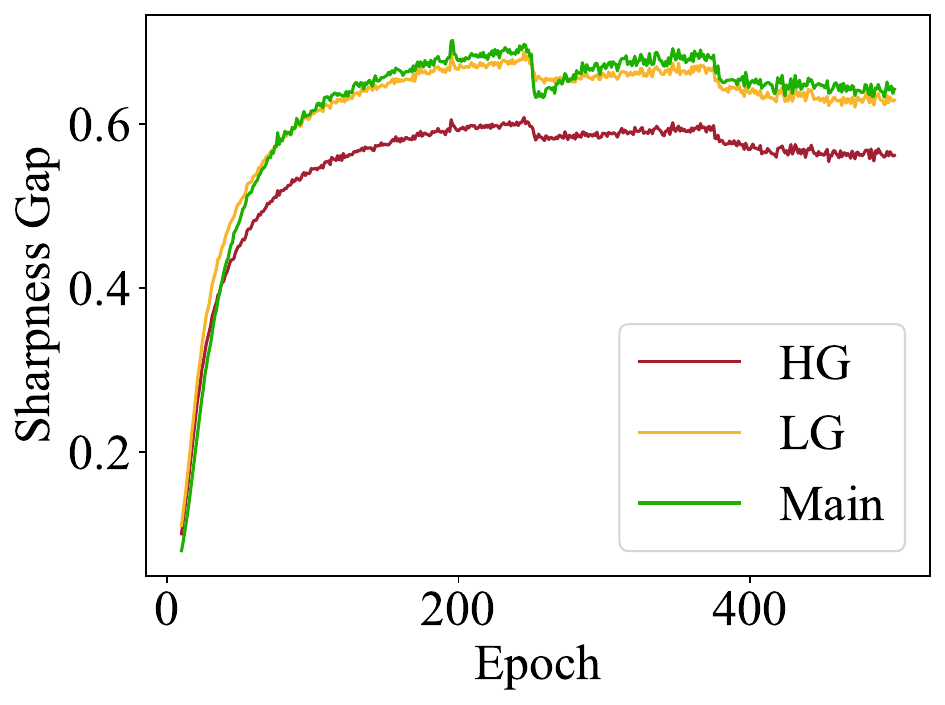}}
\subfigure[MobileNetV3]{
\label{Fig.sub.3}
\includegraphics[width=0.47\linewidth]{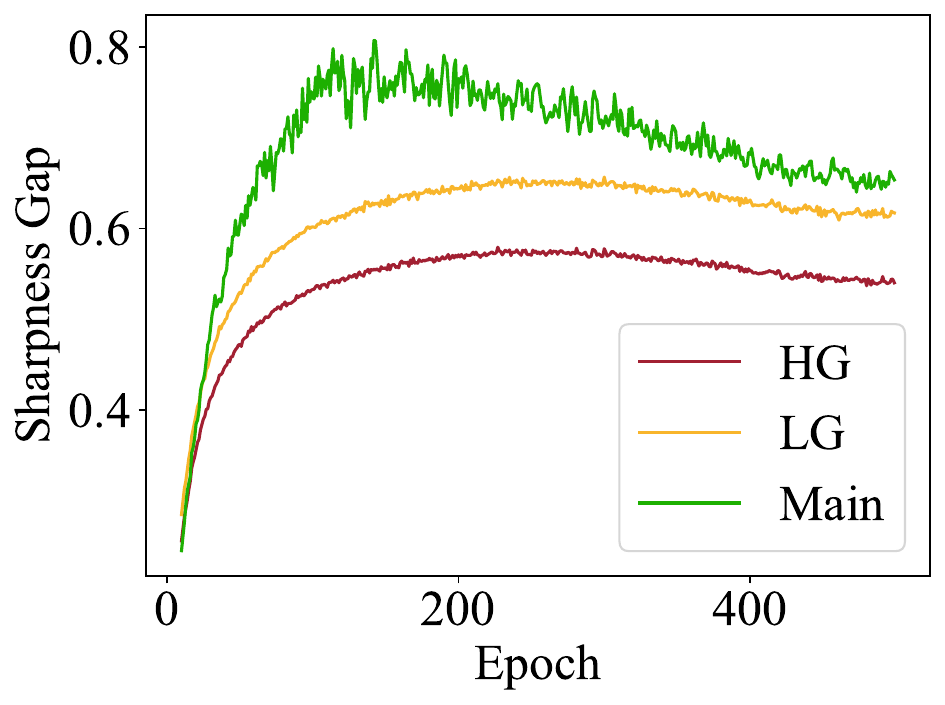}}
\caption{The capacity gap (measured by sharpness gap~\cite{guo2022reducing, rao2022parameter}) of different supervision \wl{strategies (i.e., obtaining knowledge from Largest Group (LG), Main Net (Main), and Hierarchical Group (HG)}) during the training process of (a) ResNet-50, (b) MobileNetV3.}
\label{fig:capacity gap}
\end{figure}

To measure the effect of each component, 
we remove the components of GKT one by one on WRN28-10. The results are shown in Table~\ref{table:ablation}. The first line is the performance of GKT, which is the best in the table. And it is observed that with the removal of each component, the performance is inferior step-by-step, which can prove the separate effects of each component. \tsjadd{More experiments comparing our components with other na\"ive methods are shown in \textbf{Appendix D}}.     
 

\subsection{Downstream Tasks}
\tsjadd{To verify the generalization of GKT, we finetune the ImageNet pretrained ResNet-50 of GKT and baseline on downstream tasks with three well-known frameworks including Faster R-CNN~\cite{ren2015faster}, Mask R-CNN~\cite{he2017mask}, Panoptic FPN~\cite{kirillov2019panoptic} on COCO2017~\cite{lin2014microsoft}. The results are shown in Table~\ref{table:downstream tasks}, and GKT consistently obtains improvement on object detection (e.g., +0.7\% Det mAP on Faster R-CNN), instance segmentation (+0.8\% Seg mAP on Mask R-CNN), and panoptic segmentation (+0.62\% SQ on Panoptic FPN). It demonstrates that GKT can facilitate networks to learn more general representations and benefit different tasks.}
\subsection{Investigation Experiments}
\noindent\textbf{Investigation 1: Do we need to train every subnet well?}\label{sec:train large}
ST reveals the loafing issue and proposes to train each subnet equally. However, according to ~\cite{barzilai2022kernel}, the weights of subnets contributing to the main net performance are not the same. Inspired by this, we explore the influence of different subnets on the main net. Specifically, we first divide the sampling space into several mini-spaces by the layer number of the subnet. Then we conduct ST or GKT following the uniform rule (UR)~\cite{yestimulative} on these mini-spaces and the full space, compared with following the exponential decay rule (EDR) on the full space.
As shown in Table~\ref{table:mini-spaces}, GKT performs better on larger mini-spaces, and both GKT and ST perform well on relatively large mini-spaces. Besides, following EDR, GKT and ST both achieve the best performance. These prove that we should pay more attention to training relatively large subnets.      

\noindent\textbf{Investigation 2: Does our method reduce the capacity gap?} An important purpose of GKT is to reduce the capacity gap during subnet training. To verify these,
we record the parameters of subnets in different groups in GKT \tsj{on WRN28-10}. We equally and linearly divide the parameters range into three parts including tiny part (7.4$\sim$17.17M), medium part (17.17$\sim$26.85M) and large part (26.85$\sim$36.54M), and use the recorded parameters of each group to compute the sampling ratio and parameter expectation. The results are shown in Table~\ref{table:subnet parameters}. We can observe that GKT is more inclined to sample larger subnets for larger groups, and the parameter expectation of different groups is hierarchical. Further, we quantify the capacity gap by measuring the sharpness gap, which is \bp{a} commonly used metric to represent the capacity gap~\cite{guo2022reducing, rao2022parameter}. As shown in Figure~\ref{fig:capacity gap}, compared with directly transferring knowledge from the main net or largest group, GKT can significantly and consistently obtain the lowest sharpness gap. 
\section{Conclusion}
\par In this work, from the unraveled view of residual networks, we observe that all the subnets with different capacities are provided with the same supervision in previous methods, leading to a serious capacity gap and lack of knowledge. To solve these issues, we identify the hierarchical subnet group knowledge inspired by the sociology field, and propose a novel group knowledge based training (GKT) framework to boost residual networks effectively and efficiently. Besides, we find training relatively large subnets benefit the main net more, which guides our design of the subnet sampling strategy. Comprehensive experiments on multiple tasks and networks verify GKT's generalization and superiority.

\section{Acknowledgments}
This work is supported by National Natural Science Foundation of China (No. U1909207 and 62071127), National Key Research and Development Program of China (No. 2022ZD0160100), Shanghai Natural Science Foundation (No. 23ZR1402900), and Zhejiang Lab Project (No. 2021KH0AB05).

\bibliography{aaai24}
\section*{Appendix A: Details of experiments}
In this section, we report the reproducibility details of experiments in the manuscript. All the training time is recorded on Tesla V100 and converted to GPU hours. To guarantee a fair comparison, we verify all methods with the same data augmentations, optimizer, training settings, and fine-tuned hyper-parameters. For the additional experiments presented in the supplementary materials, we use the same training settings as the manuscript. The details are as follows.
\subsection{A1. CIFAR-10/100 implementation details}
As a classical image classification dataset, CIFAR-10/100~\cite{krizhevsky2009learning} contains 50,000 training images and 10,000 testing images with 10/100 categories. For \textbf{WRN} families, we follow the training setting in~\cite{zagoruyko2016wide}, and adopt SGD as the optimizer. We train WRN for 200 epochs with a batch size of 128. And the initial learning rate is 0.1 with cosine decay schedule. The weight decay is 0.0005. For \textbf{ResNet} families and MobileNetV3, we follow the training setting in ~\cite{yestimulative}. SGD is used as the optimizer. We train ResNet and MobileNetV3 for 500 epochs with a batch size of 64. And the initial learning rate is 0.05 with cosine decay schedule. The weight decay is 0.0003. \tsjadd{For \textbf{transformers}, we follow training setting in~\cite{lee2021vision}. We optimize transformers for 100 epochs with AdamW~\cite{kingma2014adam}. The initial learning rate is 0.001. The weight decay is 0.05, and the batch size is 128. We set the warm-up to 10.} For the data augmentation, in the experiments of CNNs, we follow the setting in ~\cite{sahnicompofa} and utilize random scale transformation~\cite{yang2020gradaug}; in the experiments of transformers, we follow the setting in~\cite{lee2021vision}. For the hyper-parameters of GKT, we set \tsj{loss balanced coefficient $\beta=2$, EMA coefficient $\alpha=0.5$, the number of groups $M=3$, and the initial number $q=0.2$ of exponential decay rule.}   

\subsection{A2. Tiny ImageNet implementation details}
Tiny ImageNet is a subset of ImageNet~\cite{deng2009imagenet}. Tiny ImageNet contains 100,000 training images and 10,000 testing images with 200 categories. For \textbf{WRN} families, we follow the training setting in ~\cite{shen2022self}. We train WRN for 200 epochs with a batch size of 128. And the initial learning rate is 0.2 with a decay factor of 10\% at $100$-th, $150$-th epoch. The weight decay is 0.0001. For \textbf{ResNet} families and MobileNetV3, We train them for 500 epochs with a batch size of 64. And the initial learning rate is 0.1 with a decay factor of 10\% at $250$-th, $375$-th epoch. The weight decay is 0.0003. \tsjadd{For \textbf{transformers}, we use the same setting as CIFAR-10/100 stated above.} For the data augmentation, we follow the setting in ~\cite{shen2022self} and also utilize random scale transformation~\cite{yang2020gradaug}. For the hyper-parameters of GKT, we set loss balanced coefficient $\beta=2$, EMA coefficient $\alpha=0.5$, the number of groups $M=3$, and the initial number $q=0.2$ of exponential decay rule. 

\subsection{A3. ImageNet implementation details}
ImageNet~\cite{deng2009imagenet} is a large scale dataset consisting of 1.2 million training images and 50,000 validation images with 1000 categories. For CNNs, We follow the commonly used data augmentations as done in \cite{szegedy2015going, huang2017densely}. We employ SGD as the optimizer and train ResNet families for 200 epochs with a batch size of 512. The learning rate is 0.2 with a cosine decay schedule. The weight decay is 0.0001. \tsjadd{For the transformers, we follow the setting and data augmentation in~\cite{liu2021swin}. We employ SGD as the optimizer and train transformers for 300 epochs with a batch size of 1024. The learning rate is 0.001 with a cosine decay schedule and the weight decay is 0.05.} For the hyper-parameters of GKT, we set loss balanced coefficient $\beta=0.5$, EMA coefficient $\alpha=0.5$, the number of groups $M=3$, and the initial number $q=0.1$ of exponential decay rule.

\subsection{A4. Downstream tasks implementation details}
\tsjadd{COCO~\cite{lin2014microsoft} is a challenge and large-scale dataset for different tasks including object detection, instance segmentation, panoptic segmentation and etc.. There are 164,000 images and 879,000 annotations from 80 categories of different tasks. In our manuscript, we utilize the 2017 version of it. In the experiments of downstream tasks, we replace the backbone of Faster R-CNN~\cite{ren2015faster}, Mask R-CNN~\cite{he2017mask} and Panoptic FPN~\cite{kirillov2019panoptic} with pretrained ResNet-50 on ImageNet. For the detailed implementation, in all tasks, we follow the default settings in MMDetection~\cite{chen2019mmdetection} and adopt the standard 1× training schedule. Specifically, we train each framework for 12 epochs with a batch size of 16 with AdamW~\cite{kingma2014adam}. The initial learning rate is 0.0001 with a decay factor of 10\% at 8-th and 12-th epoch. The weight decay is 0.1.} 

\section*{Appendix B: More illustration of GKT}
\subsection{B1. Pseudo code of GKT}
The pseudo code of GKT is shown in Alg.~\ref{alg:gkt}. Although the basic symbols and notations have been introduced in the manuscript, we show them again to understand the procedure of GKT more easily. 
\par For convenience, we select image classification as the task. \tsj{Specifically, $c$ denotes the number of input channels}, $h$ and $w$ denote the image height and width respectively, $k$ is the number of classes, and $b$ is the batch size. $J$ is the total training steps. Let $x \in \mathbb{R}^{b \times c \times h \times w}$ and $y \in \mathbb{R}^{b \times k}$ denote the mini-batch samples and their ground truth labels in the training dataset. We denote $N$ as the number of all training images. The indices of images in the training dataset range from $1$ to $N$, and we denote $I \in \mathbb{N}^b$ as the indices of mini-batch $x$. $p \in \mathbb{R}^{b \times k}$ is the network output after softmax activation, which is called logits in the following. 
For a given residual network, we denote $\mathcal{N}_m$ as the main net, $\mathcal{N}_s$ as the subnet that shares weights with $\mathcal{N}_m$. The weights of the main net and subnet are denoted as $\theta _ {\mathcal{N}_m}$ and $\theta _ {\mathcal{N}_s}$ respectively. We denote $\mathcal{K}_t \in \mathbb{R}^{N \times k}$ as the $t$-th group knowledge and $\mathcal{K}_t^I \in \mathbb{R}^{b \times k}$ as the corresponding group knowledge queried from $\mathcal{K}_t$ by indices $I$. $CE(.)$ and $KL(.)$ denote the loss functions of standard cross entropy and Kullback-Leible divergence. Besides, in the pseudo-code, we denote ``=" as the assignment operator and ``==" as the equal comparison operator. In addition, ``$exit$" represents that the variable has been initialized. 
\vspace{-5pt} 
\begin{algorithm}[ht]
\caption{Group Knowledge based Training}
\label{alg:gkt}   
\begin{algorithmic}[1] 
\REQUIRE ~~\\ 
Residual main net $\mathcal{N}_m$; Total training steps $J$; Random sampling $\pi$ with exponential decay rule; Loss balanced coefficient $\beta$ and EMA coefficient $\alpha$; Input $x$ and ground truth $y$ of each minibatch; Number of groups $M$.
\STATE Construct the main net $\mathcal{N}_m$ and initialize the main network weights $\theta_{\mathcal{N}_m}$; Initialize serial number of group $t=1$.
\label{code:gkt }
\STATE For each $j\in \left[ 1, J \right]$ do
\STATE ~~~Sample a \tsj{random} subnet \tsj{and $I$-th mini-batch $x$.}:

~~~~~~if $t==1$: $\mathcal{N}_{s} = \pi(\mathcal{N}_{m})$ else: $\mathcal{N}_{s} = \pi(\mathcal{N}_{s})$

\STATE ~~~Subnet forwards:

~~~~~~$p = \mathcal{N}_{s}(\theta_{\mathcal{N}_{s}} ,x) $
\STATE ~~~\tsj{Obtain} the cross entropy loss:

~~~~~~$\mathcal{L}_{CE} = CE(p,y)$
\STATE ~~~Obtain group knowledge supervision and \tsj{obtain} the Kullback-Leible divergence loss:

~~~~~~if $t==1$:

~~~~~~~~~if $\tsj{\mathcal{K}_{1}^I} \; exits$: $\mathcal{L}_{GK} = KL(\mathcal{K}_{1}^I,p)$ else: $\mathcal{L}_{GK} = 0$

~~~~~~else: 

~~~~~~~~~if $\mathcal{K}_{t-1}^I exits$:$\mathcal{L}_{GK} = KL(\mathcal{K}_{t-1}^I,p)$ else: $\mathcal{L}_{GK}=0$

\STATE ~~~Compute the final loss:

~~~~~~$\mathcal{L}_{GKT} = \mathcal{L}_{CE} + \beta \mathcal{L}_{GK} $ 
\STATE ~~~ Update the corresponding group knowledge:

~~~~~~if $\mathcal{K}_{t}^I \; exits$: $\mathcal{K}_t^I = \alpha p + (1-\alpha) \mathcal{K}_t^I $ else: $\mathcal{K}_t^I = p$
\STATE ~~~Backward and update network weights $\theta_{\mathcal{N}_{s}}$ by descending $\nabla_{\theta_{\mathcal{N}_{s}}} \mathcal{L}_{GKT}$
\STATE ~~~Judge whether to enter the next loop and update $t$: 

~~~~~~if $t<M$: $t=t+1$ else: $t=1$
\STATE End.
\end{algorithmic} 
\end{algorithm} 
\vspace{-4pt}

\subsection{B2. Illustration of SIS sampling} \tsjadd{For accessible understanding, Figure~\ref{fig:SIS} gives the illustration of SIS sampling when there are three stages. In the first step, SIS sampling starts from the main net and drops some layers to obtain the sub-net. In the second step, the sub-net drops some layers and obtains the sub-sub-net. In the following steps, the procedure will continue until the number of steps reaches the group number and go to the next loop, which is shown in Algorithm~\ref{alg:gkt}. By adopting SIS sampling, we can build a succession relation among different sampled subnets and naturally divide all subnets into hierarchical subnet groups, which benefits transferring knowledge to subnets of different sizes.} 

\section*{Appendix C: Replicability of GKT}
To verify the replicability of GKT, we conduct experiments with two models, i.e. WRN28-10 and ResNet-50, on two datasets, i.e. CIFAR-100 and Tiny ImageNet. Each setting is run five times \tsj{under different random seeds}. We report the top-1 accuracy of each experiment and the statistics of each setting, including the average value and standard deviation. As shown in Table~\ref{tab:replicability}, with a small standard deviation, GKT can consistently and robustly achieve outstanding performance on different models and datasets.
\begin{table*}[h]
\centering
\begin{tabular}{l|llllll|l}
\hline
Dataset                        & Model     & \#1   & \#2   & \#3   & \#4   & \#5   & Statistic       \\ \hline
\multirow{2}{*}{CIFAR-100}     & WRN28-10  &  84.38 & 84.25 & 84.35 & 84.68 & 84.42 & 84.41±0.16\\
                               & ResNet-50 &  81.73 & 81.84 & 81.60 & 81.81 & 81.78 & 81.75±0.09       \\ \hline
\multirow{2}{*}{TinyImageNet} & WRN28-10  & 65.49 & 65.48 & 65.18 & 65.79 & 65.48 & 65.48±0.22 \\
                               & ResNet-50 & 67.96 & 68.16   & 67.86   & 68.08   & 68.12 & 68.03±0.12 \\ \hline
\end{tabular}
\vspace{3pt}
\caption{The top-1 accuracy of the main net of GKT with different experimental settings. Each setting is run five times \tsj{under different random seeds}.}
\label{tab:replicability} 
\end{table*}
\section*{Appendix D: More ablation and investigation experiments}
\subsection{D1. Ablation experiments of hyper-parameters}
\begin{figure*}[t]
\vspace{-10pt}
\centering  
\subfigure[Loss balanced coefficient $\beta$]{
\label{Fig.sub.1}
\includegraphics[width=0.47\linewidth]{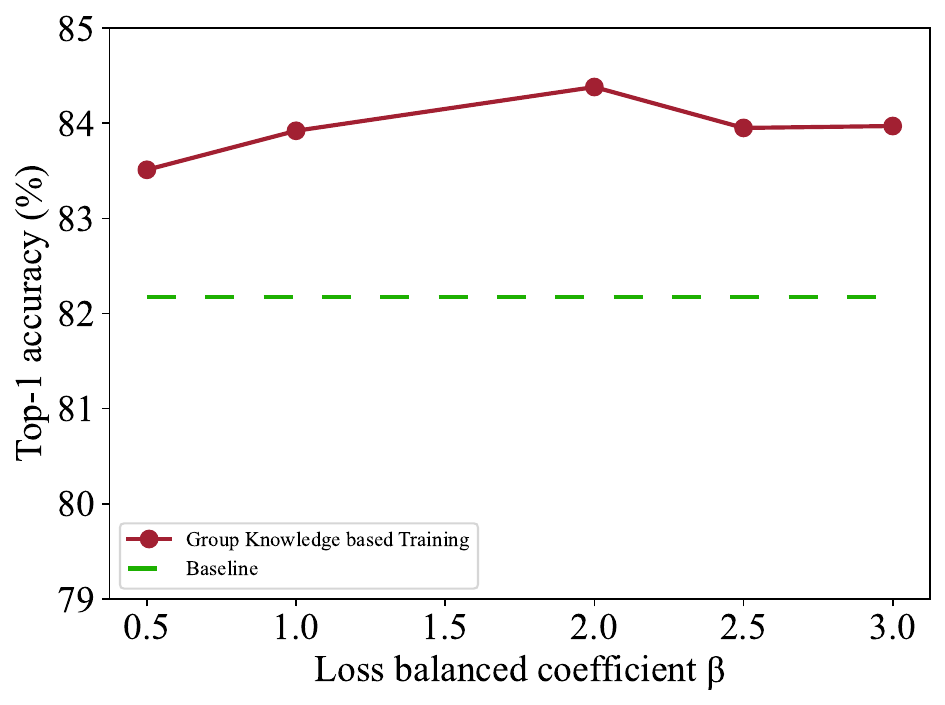}}
\subfigure[EMA coefficient $\alpha$]{
\label{Fig.sub.2}
\includegraphics[width=0.47\linewidth]{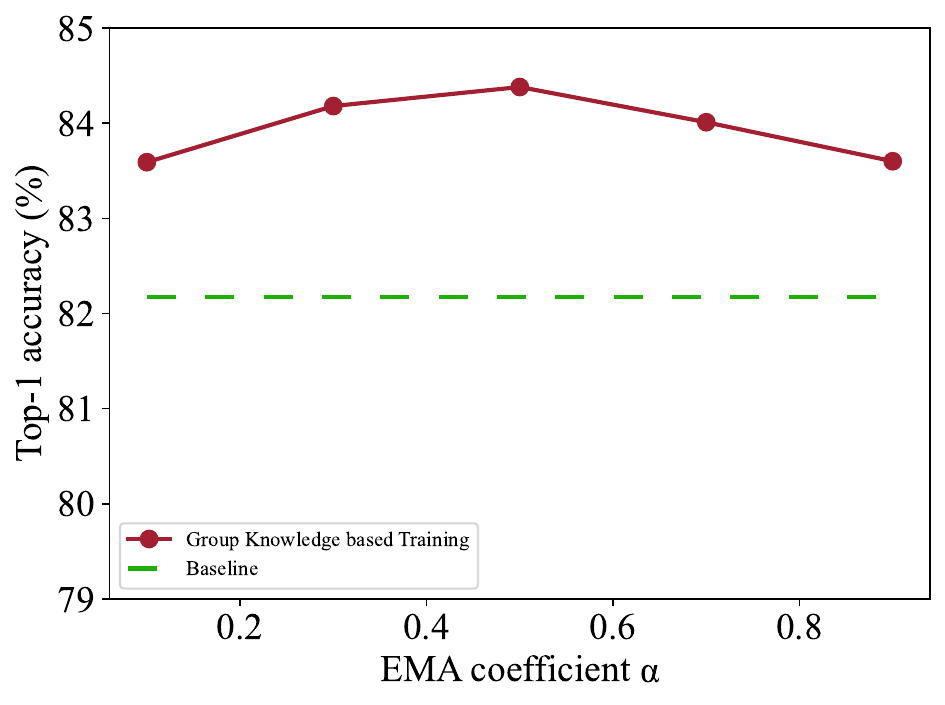}}

\subfigure[The number of groups $M$]{
\label{Fig.sub.3}
\includegraphics[width=0.47\linewidth]{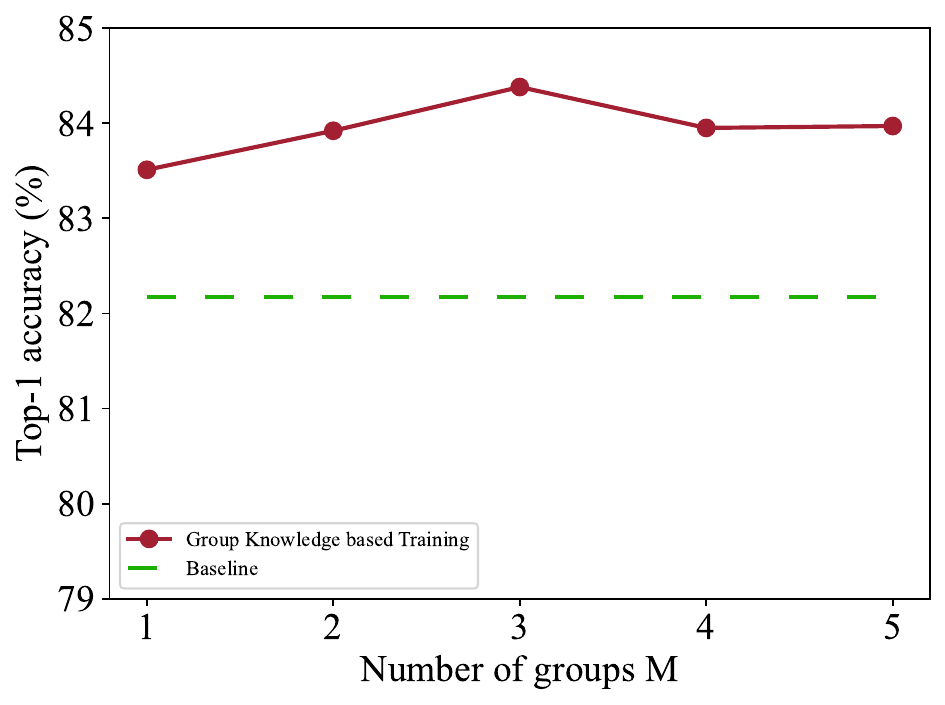}}
\subfigure[The initial number $q$ of exponential decay rule]{
\label{Fig.sub.4}
\includegraphics[width=0.47\linewidth]{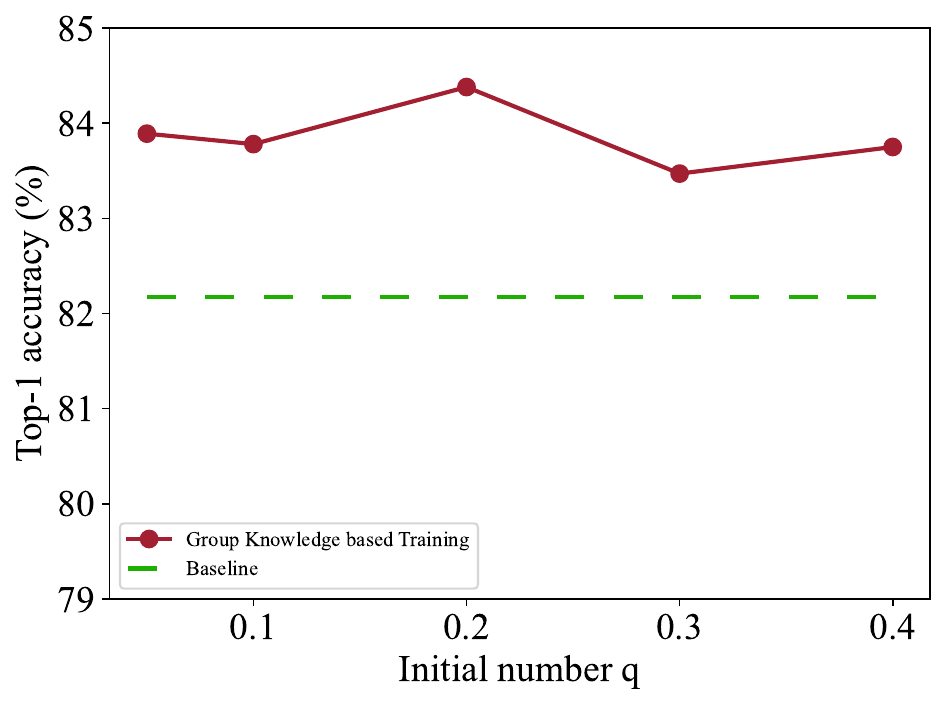}}
\caption{The ablation experiment results of varying hyper-parameters including (a) loss balanced coefficient $\beta$, (b) EMA coefficient $\alpha$, (c) number of groups $M$, and (d) the initial number $q$ of exponential decay rule on WRN28-10 and CIFAR-100.}
\label{fig:ablation}
\vspace{3pt}
\end{figure*}

There are four hyper-parameters in GKT, including the loss balanced coefficient $\beta$, EMA coefficient $\alpha$, the number of groups $M$, and the initial number $q$ of exponential decay rule. To study the effects of hyper-parameters, we conduct ablation experiments with WRN28-10 on CIFAR-100. After a primary grid search, we set $\beta=2, \alpha=0.5, M=3, q=0.2$. For studying the effect of each hyper-parameter on the results, we keep the other 3 hyper-parameters \textcolor{black}{fixed}. As shown in Figure~\ref{fig:ablation}, the performance of GKT is not sensitive to \tsj{these four} hyper-parameters. Besides, GKT can consistently surpass the baseline \textcolor{black}{by a considerable margin}, i.e. the standard training strategy.

\subsection{D2. SIS Sampling v.s. Others}
To verify the superiority of 
subnet-in-subnet (SIS) sampling, we intuitively design two na\"ive group division methods, 
namely FLOPs based division and parameters based division. Specifically, we linearly and equidistantly divide all the possibly sampled subnets into several groups based on FLOPs or parameters in advance and randomly sample a subnet during training. 
\par As \bp{shown in} Table~\ref{SIS vs others}, on ResNet-34 and WRN28-10, SIS sampling is significantly superior to na\"ive group division methods. We consider the superiority can be explained as the more extensive knowledge ensemble, as group knowledge can be seen as a kind of weighted knowledge of subnets belonging to the group. For each subnet group, the strict group division methods only contain relatively limited subnets. As a comparison, SIS sampling permits every subnet group to possibly contain every subnet, leading to a more extensive knowledge ensemble. In detail, the group knowledge of SIS sampling can be seen as a weighted ensemble of all sampled subnets, and for a large group the weights of large subnets are larger than tiny subnets.
\begin{table}[t]
\centering
\begin{tabular}{l|cc}
\hline
Method               & ResNet-34     & WRN28-10       \\ \hline
FLOPs based division     & 80.46         & 83.57          \\
Parameters based division & 80.48         & 83.13          \\
SIS sampling           & \textbf{81.40} & \textbf{84.38} \\ \hline
\end{tabular}
\caption{Effectiveness of different partition strategies for dividing all subnets into multiple subnet groups.}
\label{SIS vs others}
\end{table}

\subsection{D3. Hierarchical Group Knowledge Transfer v.s. Others}
\tsj{To verify the effectiveness of hierarchical group knowledge transfer (HGKT), we compare it with obtaining knowledge form 
Largest Group (LG), Self Group (SG), and Average Group (AG), on ResNet-34 and WRN28-10. As shown in Table~\ref{table:transfer}, HGKT can consistently achieve the most excellent result. The reason is that, LG only transfers knowledge from the largest group and has more capacity gap; SG only utilizes knowledge from the self group and lacks knowledge with higher quality; AG averages the knowledge of all groups and ignores the gap between groups. These results further verify that suitable supervision should have less capacity gap and abundant knowledge.}  
 \begin{table}[]
\centering
\begin{tabular}{l|cc}
\hline
Method                 & Resnet-34     & WRN28-10       \\ \hline
LG     & 80.12         & 83.80          \\
SG    & 80.63         & 83.79          \\
AG       & 80.61         & 83.90          \\ 
HG & \textbf{81.40} & \textbf{84.38} \\\hline
\end{tabular}
 \caption{Effectiveness of different supervision strategies. \tsj{For a full comparison, we obtain knowledge from the Largest Group (LG), Self Group (SG), Average Group (AG), and the proposed Hierarchical Group (HG) to supervise diverse subnets.}}
 \label{table:transfer}
\end{table}

\subsection{D4. Exponential Decay Rule v.s. Others}
For a fair comparison, we only replace the exponential decay rule in our GKT framework with commonly-used rules, such as uniform rule~\cite{yestimulative} and linear decay rule~\cite{huang2016deep}.
Since linear decay rule~\cite{huang2016deep} is not stage-wise, 
we also design its stage-wise variant, namely linear decay (stage-wise) rule. As shown in Table~\ref{table:sampling strategies}, \tsj{on WRN28-10}, the proposed exponential decay rule achieves the best performance. Besides, compared with the uniform rule, the exponential decay rule is prone to sampling larger subnets which can more significantly benefit main net performance.

\subsection{D5. Large subnets v.s. all subnets}
Further, we utilize different training methods and test the performance of all sampled subnets and relatively large subnets (with 9-12 residual blocks). 
We additionally design a variant of GKT by training subnets and main net alternately, vividly named GKT-ABAC. The results are shown in Table~\ref{table:influence of different subnet training methods}. Similar to~\cite{mok2022demystifying}, we utilize Kendall’s Tau rank correlation to quantify the relationship. 
The Kendall's Tau rank correlation is 0 between all subnets and main net performance, and 0.67 between large subnets and main net performance, which means improving large subnets' performance is more likely to improve main net performance compared with improving all subnets' performance.
\begin{table*}[t]
\centering
\begin{tabular}{l|c|c|c}
\hline
Method                & Top-1 Acc(\%)  & Subnet Top-1 Acc(\%) & Large subnet Top-1 Acc(\%) \\ \hline
Baseline              & 82.17          & 49.12±18.17          & 72.19±9.96                 \\
GKT-UR & 82.21          & 80.08±1.09           & 80.99±0.05                 \\
GKT-ABAC             & 82.73          & \textbf{81.21±1.56}  & 82.60±0.1                  \\
GKT                  & \textbf{84.38} & 72.36±10.73          & \textbf{83.61±0.53}        \\ \hline
\end{tabular}
\caption{Influence of different subnet training methods on the final performance, the averaged performance of all subnets, and the averaged performance of large subnets. GKT-ABAC is a variant of GKT by training subnets and main net alternately. GKT-Uniform Rule is GKT with the uniform rule, i.e. sampling all subnets with equal probability.}
\label{table:influence of different subnet training methods}
\end{table*}
\begin{table}[]
\centering
\begin{tabular}{l|c}
\hline
Sampling Strategy            & Top-1 Acc(\%)  \\ \hline
Uniform Rule             & 81.06          \\
Linear Decay Rule & 83.35          \\
Linear Decay (stage-wise) Rule  & 83.48          \\
Exponential Decay Rule             & \textbf{84.38} \\ \hline
\end{tabular}
\caption{Effectiveness of \tsj{GKT equipped with different sampling rules}.}
\label{table:sampling strategies}
\end{table}

\section*{Appendix E: Capacity gap on TinyImageNet}
We further verify that GKT can always reduce the capacity gap compared with obtaining supervision from the main net \tsj{(i.e., Main)} or largest group \tsj{(i.e., LG)} on Tiny ImageNet. As shown in Figure~\ref{fig:capacity gap}, based on either ResNet-50 or MobileNetV3, the proposed hierarchical group knowledge transfer, i.e. the HG in the figure, can consistently reduce the capacity gap \textcolor{black}{more significantly} than the other two supervision strategies.
\begin{figure}[ht]
\vspace{-10pt}
\centering  
\subfigure[ResNet-50]{
\label{Fig.sub.2}
\includegraphics[width=0.47\linewidth]{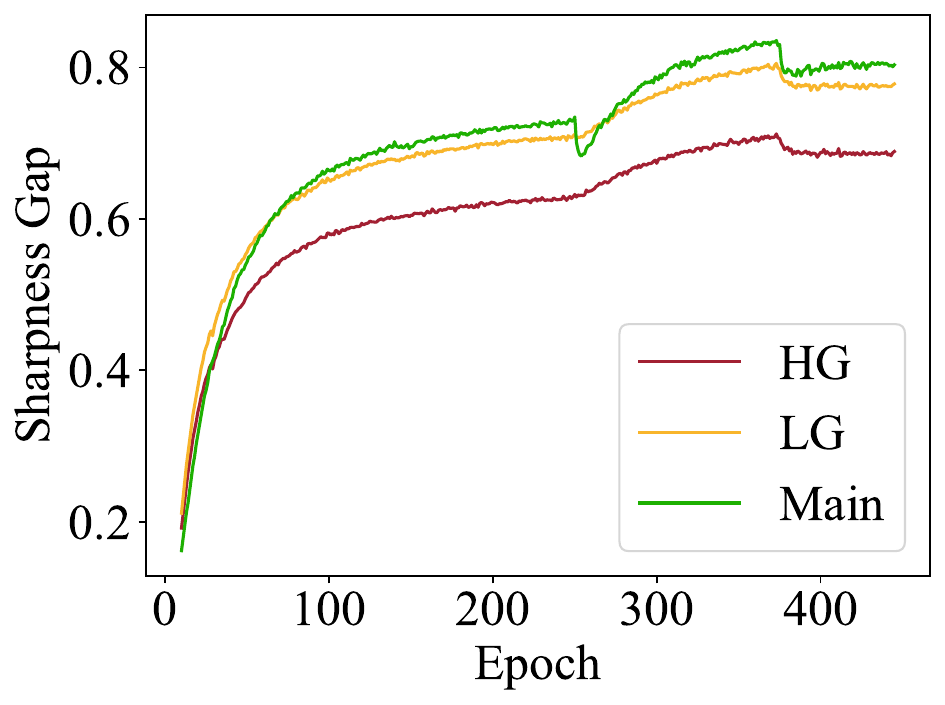}}
\subfigure[MobileNetV3]{
\label{Fig.sub.3}
\includegraphics[width=0.47\linewidth]{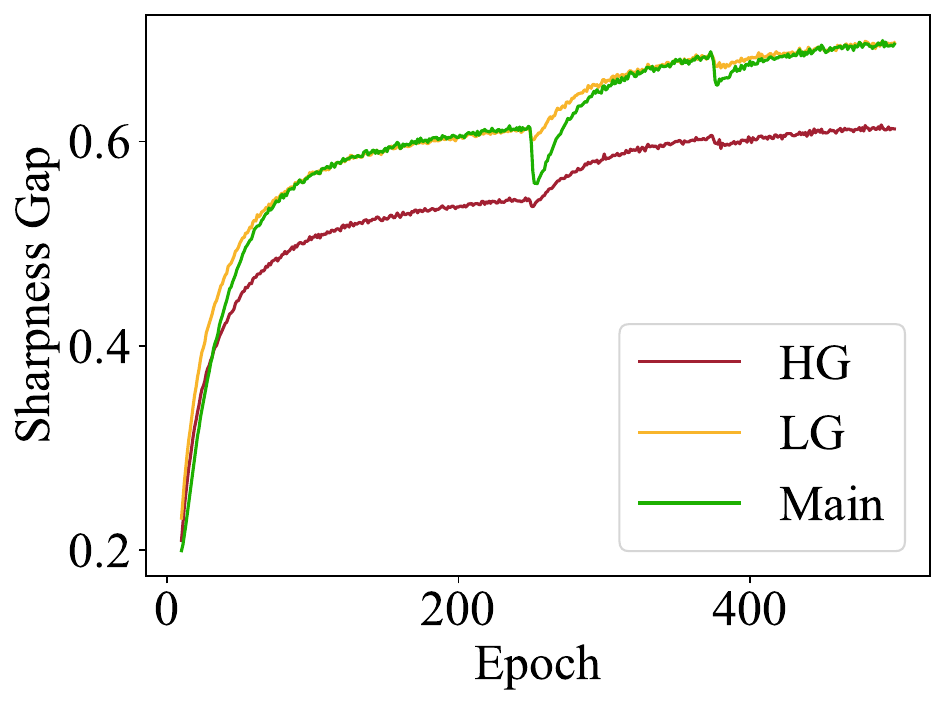}}
\caption{The capacity gap (measured by sharpness gap~\cite{guo2022reducing, rao2022parameter}) of different supervision strategies (i.e., obtaining knowledge from Largest Group (LG), Main Net (Main), and Hierarchical Group (HG)) during the training process of (a) ResNet-50, (b) MobileNetV3 on TinyImageNet.}
\label{fig:capacity gap}
\vspace{3pt}
\end{figure}
\section*{Appendix F: Efficiency and effectiveness of GKT}
To further verify the efficiency and effectiveness of GKT, we additionally report the training cost(GPU hours and memory) and Top-1 accuracy(\%) on CIFAR-100. \tsjadd{The results are shown in Table~\ref{table:memory_and_time}. Compared with baseline and other methods, GKT has the least training time and obtains the best Top-1 accuracy. As shown in Figure~\ref{fig:compare_ideas}, the results on CIFAR-100 are similar to that on Tiny ImageNet (shown in Figure~2 of the manuscript). Thus we conclude that GKT can achieve optimal efficiency and performance trade-offs on multiple datasets and networks. The explanation for the efficiency is that \textbf{only one subnet with lower FLOPs forwards and backwards at each iteration.} Therefore, it costs lower training time when compared with the main net. The extra memory required is negligible, because the group knowledge refers to logits of a sample from one subnet group, which are 100/1000 elements for CIFAR100/ImageNet-1K per image/sample in each iteration.}

\begin{table*}[ht]
\centering
\begin{tabular}{l|l|lllllll|l}
\hline
              & Baseline & StoDepth  & ONE     & CS-KD  & PS-KD  & LWR    & DLB    & ST     & GKT            \\ \hline
Top-1 Acc(\%) & 82.17    & 82.75  &   83.02      & 81.25  & 82.53  & 82.00  & 81.35  & 82.84  & \textbf{84.38} \\
GPU hours     & 3.6h     & 3.4h   &  8.1h       & 3.8h   & 3.9h   & 3.8h   & 6.8h   & 7.6h   & \textbf{3.3h}  \\
Memory(MB)    & 4459MB   & \textbf{4320MB} & 7011MB & 5957MB & 5433MB & 4623MB & 7031MB & 4581MB & 4486MB         \\ \hline
\end{tabular}
\caption{The training cost(GPU hours and memory) and Top-1 accuracy(\%) of different methods on WRN28-10 on CIFAR-100. GKT can obtain the optimal trade-off of efficiency and effectiveness with marginal memory increase.}
\label{table:memory_and_time}
\end{table*}

\begin{figure}[ht]
  \centering
  \includegraphics[width=\linewidth]{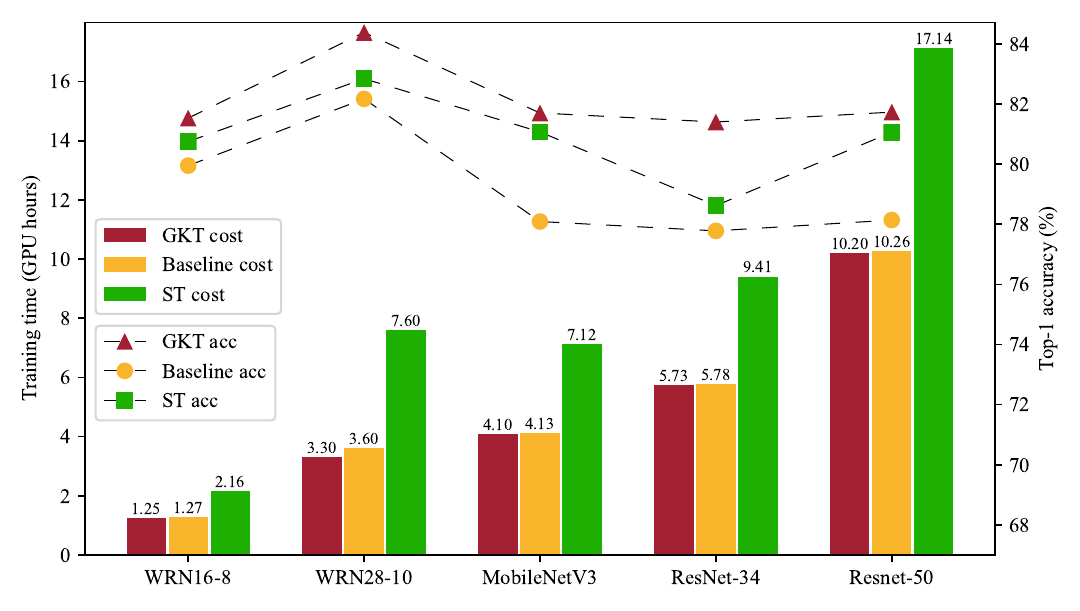}
  \vspace{-6pt}
  \caption{The efficiency and effectiveness of GTK on CIFAR-100}
  \label{fig:compare_ideas}
  \vspace{-6pt}
\end{figure}

\section*{Appendix G: Compared with other methods}
\subsection{G1. Advantages Over Previous Subnet Training}
\par Residual networks can be seen as an implicit ensemble of relatively shallow \yep{networks}~\cite{veit2016residual}, and invariably suffer from the network loafing problem~\cite{yestimulative}. 
Previous methods \yep{have attempted to strengthen residual networks by training their subnets}. Stochastic depth (Stodepth)~\cite{huang2016deep} supervises randomly sampled subnet by ground truth via optimizing the following loss:
\begin{align}
\label{formulation:sd loss} \mathcal{L}_{stodepth} = CE(p(\theta _ {\mathcal{N}_s},x),y).
\end{align} 
And stimulative training (ST) ~\cite{yestimulative} forwards the network twice to supervise \yep{both} the main net and randomly sampled subnet via optimizing the following loss:

\begin{footnotesize} 
\begin{align}
\label{formulation:st loss} 
\mathcal{L}_{ST} = CE(p(\theta_{\mathcal{N}_m},x ),y) + \lambda KL(p(\theta_{\mathcal{N}_m},x),p(\theta_{\mathcal{N}_s} ,x)) 
\end{align}
\end{footnotesize} 

\noindent where $\lambda$ is the loss balanced coefficient.
From Equation~\ref{formulation:sd loss} and~\ref{formulation:st loss}, we can observe a mismatch that \yep{the same kind of supervision is provided for variously sized subnets, ignoring different characteristics of various subnets.
This mismatch will cause a serious capacity gap that hinders knowledge transfer}. What's more, for Stodepth, ground truth is short of knowledge, e.g., inter-class information. For ST, it needs an additional main net \yep{forwarding that} consumes extra computation resources. To overcome the dilemmas of ST and Stodepth, \yep{\textbf{we attempt to find suitable supervision for diverse subnets with abundant knowledge, less capacity gap, and no extra computation.}
Observing that there are already} different subnets and their outputs in the subnet training \yep{process, we propose to divide all subnets into hierarchical groups, aggregate subnet historic outputs in the same group, and use it as suitable supervision for the neighboring group.} Interestingly, in the field of sociology and management science, \yep{due to} the diversity of members, transferring knowledge to individuals is also a \yep{challenging} task. And a common solution is to collect the individual knowledge of the same 
producing group and transfer it to the neighboring group~\cite{kane2005knowledge, erden2008quality}. We believe it is heuristically similar to our method and is a vivid analogy helping our method to be comprehended easily. Thus, \yep{we name our} method group knowledge based training (GKT).  
\subsection{G2. Visual comparison with other methods}
\begin{figure*}[t]
  \centering
  \includegraphics[width=\linewidth]{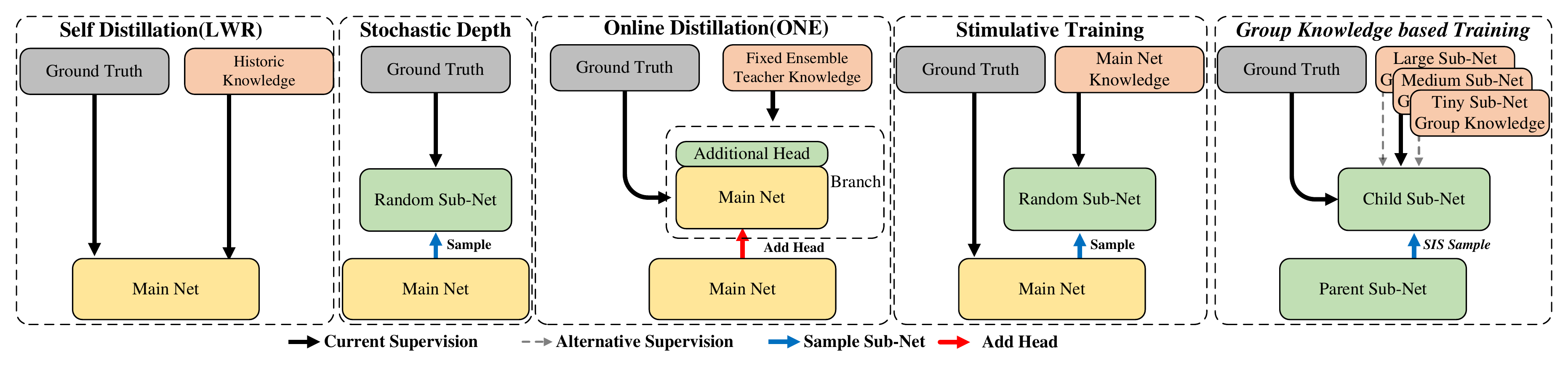}
  \caption{Illustration of different methods including Self Distillation(LWR~\cite{deng2021learning}), Stochastic Depth~\cite{huang2016deep}, Online Distillation(ONE~\cite{zhu2018knowledge}), Stumulative Training~\cite{yestimulative} and the proposed GKT scheme. Different from other works that use the same kind of supervision/knowledge for all subnets, we divide all subnets into hierarchical subnet groups by subnet-in-subnet (SIS) sampling and use the aggregated knowledge of the neighboring larger group to supervise the current subnet without any extra architectures.}
  \label{fig:compare_ideas_revised}
\end{figure*}

\begin{figure}[t]
  \centering
  \includegraphics[width=0.98\linewidth]{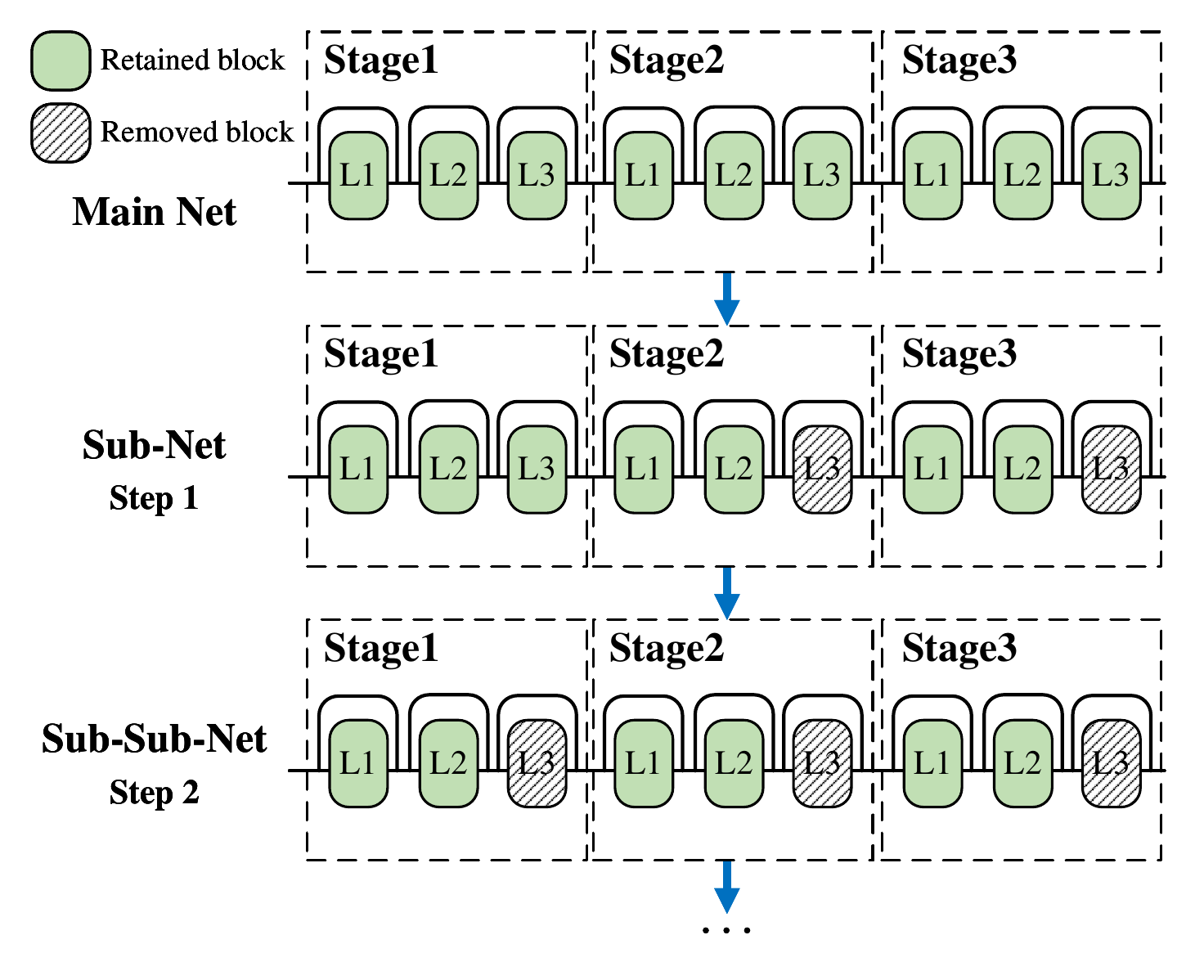}
  \caption{Illustration of the subnet-in-subnet (SIS) sampling, which can naturally divide all subnets into hierarchical subnet groups of different sizes. We suppose there are three stages in the main net.}
  \label{fig:SIS}
\end{figure}

\tsjadd{For a more intuitive understanding of the proposed GKT, Figure~\ref{fig:compare_ideas_revised} shows the illustrations of GKT and other related methods. It can be observed that although almost all other methods introduce additional supervision, they use the same kind of supervision/knowledge for branch/subnet/main net without selecting. Different from them, GKT can dynamically sample hierarchical subnet groups and aggregate knowledge of the neighboring larger
group to supervise the current subnet without any extra architectures. GKT aims to provide suitable and tailored supervision for each unique subnet with a lower capacity gap.}  

\end{document}